\documentclass[lettersize,journal]{IEEEtran}
\usepackage{amsmath,amsfonts}
\usepackage{algorithmic}
\usepackage{algorithm}
\usepackage{array}
\usepackage[caption=false,font=normalsize,labelfont=sf,textfont=sf]{subfig}
\usepackage{textcomp}
\usepackage{stfloats}
\usepackage{url}
\usepackage{verbatim}
\usepackage{graphicx}
\usepackage{cite}
\usepackage{amsmath}
\usepackage{amssymb}
\usepackage{mathtools}
\usepackage{amsthm}

\usepackage{booktabs}
\usepackage{multirow}
\usepackage{array}
\usepackage{xcolor}
\usepackage{colortbl}
\usepackage{hyperref}
\usepackage[textsize=tiny]{todonotes}

\usepackage[capitalize,noabbrev]{cleveref}
\usepackage{hyperref}
\begin{document}

\title{Personalized Face Privacy Protection From a Single Image}

\author{Zachary Yahn, Fatih Ilhan, Tiansheng Huang, Selim Tekin, \\ Sihao Hu, Yichang Xu, Margaret Loper, Ling Liu,~\IEEEmembership{Fellow, IEEE}
\thanks{Manuscript received April 30, 2026 \\ This work was funded by the GTRI PhD Fellowship. \\ The authors are with the Georgia Institute of Technology, Atlanta, Georgia, USA. (email: zachary.yahn@gatech.edu)}}




\maketitle

\begin{abstract}
Photos of faces uploaded online are vulnerable to malicious actors who can scrape facial images from online sources and intrude on personal privacy via unauthorized use of facial recognition models. This paper presents \textsc{FaceCloak}, a novel personalized face privacy protection system, which can generate defensive identity-specific universal face privacy masks from a single image of a user, causing facial recognition to fail. \textsc{FaceCloak} introduces a three-stage personalized face perturbation learning methodology: (1) It generates a small set of high-variety synthetic face images of a person based on a single image of the person. (2) It learns face cloaking by adding more protection to key facial-identity leakage regions through iterative perturbation generation over the small set of synthetic images, effectively shifting a user's identity embedding towards a distant anchor identity and away from a similar one. (3) It generates a personalized identity-protective mask in the form of pixel-wise cloaking, which is light-weight and can be efficiently applied to any facial image of a user while maintaining good perceptual quality. Extensive experiments on three popular face datasets across ten recognition models show the effectiveness of \textsc{FaceCloak} compared to 29 other existing representative methods. Code is available \href{https://github.com/zacharyyahn/FaceCloak}{here}.
\end{abstract}


\begin{figure*}[t]
    \centering
    \includegraphics[width=2\columnwidth]{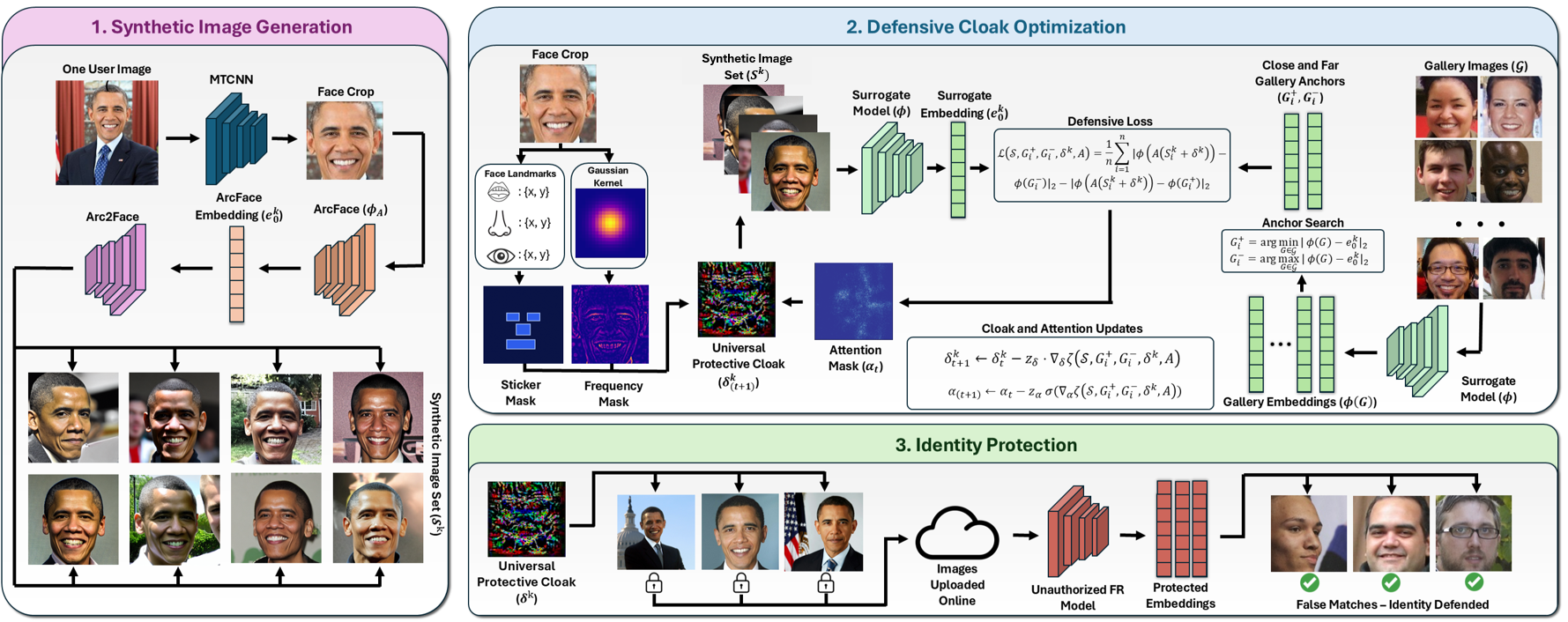}
    \caption{Overview of \textsc{FaceCloak} with three progressive stages for face privacy protection: synthetic facial image generation, defensive face cloak optimization, and face-based identity protection. A user only needs to supply a single face image of themselves and will receive an identity-specific face privacy mask that can be rapidly added to any of their face images prior to online release.}
    \label{fig:overview}
\end{figure*}
\label{sec:intro}

\section{Introduction}

Facial Recognition (FR) has long been an important concern for individual privacy. 
People wish to post their face photos online to socialize, blog, network, and connect with their colleagues, families, and friends, but they do not want unauthorized collection of their identities in large-scale databases. These databases can support infringements on real-world privacy by individuals, such as via stalking and identity fraud \cite{aisonline2021faceinus, equifax2023identityrisk}, as well as by corporations, such as via surveillance, targeted advertisements, and tracking \cite{nyt2021secretcompany}. Examples include PimEyes \cite{pimeyes2025pimeyes}, which provides a searchable database of public facial photos, and Clearview AI \cite{clearview2025clearview}, which provides facial data to governments. In light of these threats to privacy and given the prevalence of open-source facial recognition models \cite{deng2018arcface, wang2018cosface, howard2017mobilenet, boutros2022sface}, there is a pressing need to protect user images uploaded online from unwanted facial recognition while maintaining image quality.

A popular defense against unauthorized facial recognition is to subtly perturb a person's face images with strategic noise (often dubbed a `cloak') to transpose their embeddings to a different latent identity class prior to uploading. However, most existing face perturbation methods require numerous backpropagation calls on a surrogate model \cite{shan2020fawkes, cherepanova2021lowkey, Wenger2021SoKAR, chandrasekaran2020faceoff} or inference on a large generative model \cite{hu2022amtgan, Sun2024diffam, An2024SD4Privacy, shamshad2023clip2protect} for every new image, which may be prohibitively expensive for the average user. Instead of generating a new mask for each image, identity-specific (often dubbed `universal') perturbations optimize a personalized defensive perturbation once for each user \cite{zhong2022OPOM}. Once generated, such perturbations can be applied to new images of that user trivially. This presents a new set of problems: images provided by users may not be sufficiently representative, the user may not have enough images available to supply the learning process, or the privacy-minded user may not want to share many images of themselves in the first place.

With these problems in mind, we present  \textsc{FaceCloak}, a novel three-stage system which needs only one face image from a user to generate their identity-specific face privacy cloak. First, we generate a small set of diverse, high-quality synthetic images using a single seed image provided by the user. These synthetic images portray the user with varying expressions, lighting, poses, clothing, and backgrounds. Second, we optimize a personalized identity-specific privacy perturbation over the set of synthetic images to shift their embeddings towards a distant latent identity class. We constrain this optimization with a perturbation budget hyperparameter to ensure that the images perturbed with the identity-specific face privacy cloak will maintain high perceptual quality. Third, \textsc{FaceCloak} provides a personalized identity-specific face perturbation mask that can be applied to any new image of the user through the combination of diverse perturbation focusing methods, e.g., Region-Sticker, Learnable Attention, and High-Pass Mask. 

We validate our approach with experiments on three face datasets with ten facial recognition models compared with 29 existing representative identity-specific and image-specific face perturbation methods. Our results show that \textsc{FaceCloak} substantially outperforms all other identity-specific and image-specific methods in the literature. We also visually compare \textsc{FaceCloak} to examples from the literature to show that it  maintains a high perceptual quality for cloaked face images.

\section{Related Work}

\noindent
{\bf Image-Specific Perturbations.\/}
Image-specific perturbations learn a new perturbation for every image regardless of user identity~\cite{shan2020fawkes, evtimov2020foggysight, cherepanova2021lowkey, chandrasekaran2020faceoff, Wenger2021SoKAR}. Noise-based image-specific methods mainly differ by how different parts of the loss function are constructed, aiming to design an iterative method for optimizing stealthy perturbations~\cite{Yang2020tipim}. A different thread of image-specific work proposed applying makeup to faces instead of learning noise-based perturbations~\cite{yin2021advmakeup}. Some use generative adversarial networks \cite{goodfellow2014gan}, such as AMT-GAN \cite{hu2022amtgan}, while others, like DiffAm, \cite{Sun2024diffam} use a fine-tuned diffusion model \cite{Ho2020diffusion}. Similarly, CLIP2Protect \cite{shamshad2023clip2protect} uses zero-shot CLIP \cite{Radford2021clip} embeddings to generate protective makeup. These makeup-based methods achieve high protection rates, but the makeup masks are often highly visible and may be undesirable for people who do not wear makeup. 

Existing works have also attempted to combine perturbation-based and makeup-based techniques by training generative models to produce stealthy privacy masks. AdvFaces \cite{Deb2019AdvFaces} synthesizes protected face images with a GAN \cite{goodfellow2014gan}. SD4Privacy \cite{An2024SD4Privacy} uses Stable Diffusion \cite{Rombach2021stablediffusion} to generate adversarial perturbations that blend with a user's face. This achieves strong stealthiness scores but lacks the advanced protection of other methods. Adv-CPG \cite{Wang2025AdvCpg} proposes a different approach by generating portraits optimized against facial recognition; however, this significantly alters the original image.

\noindent
{\bf Identity-Specific Perturbations}.
These perturbations are learned offline once and rapidly applied online to any facial images of the user. One Person One Mask (OPOM) \cite{zhong2022OPOM} produces universal perturbations by optimizing over many images of the same user in order to ensure that the perturbation generalizes to any image of that user. AdvCloak \cite{liu2023advcloak} proposes to first learn image-specific perturbations and then combine them into identity-specific perturbations with a GAN~\cite{goodfellow2014gan}. P3-Mask \cite{chow2024chameleon} proposes an ensemble method to optimize identity-specific universal perturbations. All of these identity-specific perturbation methods require many images for each user to either train the model, as with AdvCloak, or optimize a perturbation, as with OPOM and P3-Mask. We present \textsc{FaceCloak}, which only asks each user to provide one face image and learns to generate identity-specific perturbations with high defense performance and high stealthiness. 

\section{Methodology}

\subsection{Problem Definition and Solution Overview}
Given face image $I_0^k \in \mathbb{R}^{H \times W \times C}$ for a person with identity $k$, the goal of facial privacy protection is to imperceptibly manipulate the pixels of $I_0^k$ with perturbation $\delta_0^k \in \mathbb{R}^{H \times W \times C}$ so that the embedded representation of $I_0^k$ is closer to an image of a different identity $G_n^j$ than every other image of the person's actual identity $G_n^k$ in gallery set $\mathcal{G}$. We formalize this goal as:

\begin{equation}
\begin{split}
\|\phi(I_0^k + \delta_0^k) - \phi(G_n^j)\|_2 < \|\phi(I_0^k + \delta_0^k) - \phi(G_n^k)\|_2, \\
\quad \forall G^k \in \mathcal{G}, \exists G^j \in \mathcal{G}, j \neq k
\end{split}
\end{equation}

Here $\phi: \mathbb{R}^{H \times W \times C} \to \mathbb{R}^d$ is a face image embedding function. Typically, $\delta_0^k$ is optimized to protect a single image. In universal facial privacy protection, we attempt to find an identity-specific perturbation $\delta^k$ that achieves the above condition when applied to any image of a specific user $I_n^k \in \mathcal{I}^k =\{I_0^k, I_1^k, \ldots, I_n^k\} $ where $\mathcal{I}^k$ is the set of publicly available images for a person with identity $k$.


To find $\delta^k$, we propose a methodology with three phases, as outlined in Figure \ref{fig:overview}. First, given a single image of a person who wishes to protect their facial privacy, we generate a set of synthetic images of that person. Second, we optimize $\delta^k$ over this set of synthetic images, prioritizing key regions or pixels that our method either learns or computes before iterating. Third, we keep this identity-specific $\delta^k$ so that when it is added to any image of the person it causes a facial recognition model to misidentify them.

\subsection{Synthetic Image Generation}
Previous universal facial privacy protection methods require several user images in order to optimize identity-specific perturbation $\delta^k$ \cite{chow2024chameleon, liu2023advcloak, zhang2025ctuap, zhong2022OPOM}. This is costlier for the user and, given that perturbation optimization is prohibitively expensive for a typical consumer device, it requires them to risk more privacy by transmitting multiple images to the compute server. Instead, we generate synthetic images using just one image from a user. Given user image $I_0^k$ with ArcFace \cite{deng2018arcface} embedding $e_0^k = \phi_{ArcFace}({I_0^k})$, we generate synthetic images $\mathcal{S}^k = \{S_0^k, S_1^k, ..., S_n^k\}$ with $\mathcal{S}^k = \Psi(e_0^k, n)$, where $\Psi$ is the Arc2Face \cite{papantoniou2024arc2face} synthetic image generation model and $n$ is the desired number of synthetic images. In practice, we find that small values of $n$ (e.g. 8) are sufficient to learn strong defensive perturbations due to the quality and variability in synthetic images. Arc2Face naturally generates a wide variety of synthetic user images, including different facial expressions, lighting conditions, pose angles, and clothing.

\subsection{Defensive Cloak Optimization}
We optimize $\delta^k$ over the set of synthetic user images $\mathcal{S}^k$ to achieve facial privacy protection of any real user image in $\mathcal{I}^k$. First, we identify near and far anchor images from gallery set $\mathcal{G}$ with respect to the input user image $I_0^k$ with embedding $e_0^k$: 

\begin{equation}
G^+ = \arg\min_{G \in \mathcal{G}} \|\phi(G) - e_0^k\|_2
\end{equation}
\begin{equation}
G^- = \arg\max_{G \in \mathcal{G}} \|\phi(G) - e_0^k\|_2
\end{equation}

Where $G^+$ is the nearest image in the gallery set and $G^-$ is the farthest. With these anchors we define our contrastive loss function:

\begin{equation}
\label{eq:loss}
\begin{split}
\mathcal{L}(\mathcal{S}, G^+, G^-, \delta^k, A) = \frac{1}{n} \sum_{i=1}^{n} \|\phi(A(S_i^k+\delta^k)) 
- \phi(G^-)\|_2 \\ - \|\phi(A(S_i^k+\delta^k)) - \phi(G^+)\|_2
\end{split}
\end{equation}



\begin{figure}
    \centering
    \includegraphics[width=0.9\columnwidth]{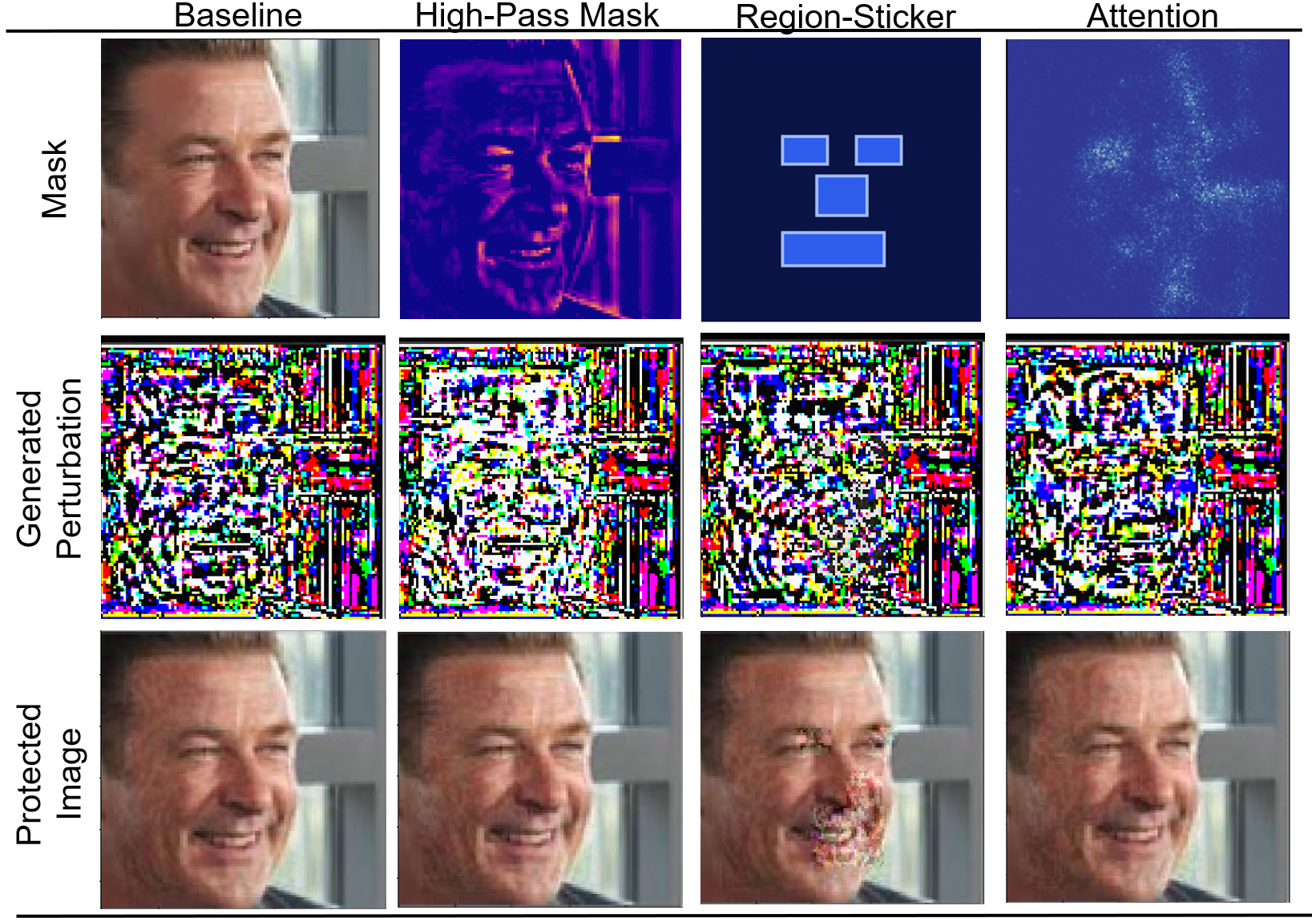}
    \caption{Visual examples of High-Pass Mask, Region-Sticker, and Learnable Attention mask focusing mechanisms. Each mechanism is designed to emphasize a different aspect of the perturbation with respect to the original image: High-Pass masks hide more noise in high-frequency areas, Region-Sticker masks emphasize perturbations on key identifying features, and Learnable Attention masks add more noise to pixels that are identified as more important to defense during optimization. We use all three in our method.}
    \label{fig:perts}
\end{figure}

In addition to optimizing our perturbation by minimizing this loss, we define three types of focusing functions $A$ that operate on the pixels of the perturbation: Region-Stickers, High-Pass Masks, and Learnable Attention denoted $A_{sticker}$, $A_{highpass}$, and $A_{attention}$, respectively. We define these functions in the following three subsections. We illustrate these and their effects on the generated perturbation in Figure \ref{fig:perts}. Our optimization methodology applies these in any combination to focus the defensive perturbation on key areas of the facial image and increase protection without sacrificing stealth.

\subsubsection{Region-Stickers} 
Intuitively, certain areas of a face are more important to facial recognition than others. We propose that these regions include a person's eyes, nose, and mouth. We hypothesize that larger perturbation magnitudes on these regions may increase protection success by blocking regions that are essential for identification. We extract landmarks $L = \{l_{\text{left eye}}, l_{\text{right eye}}, l_{\text{nose}}, l_{\text{mouth}}\}$ for these with $L = \Lambda(I_0^k)$ where $\Lambda$ is the MTCNN \cite{zhang2016mtcnn} landmark detection model from InsightFace \cite{deng2021insightface}. We then define bounding boxes around each landmark corresponding to the approximate proportional size of the respective facial feature, which we coin as ``stickers:" $B_i = \text{Box}(l_i, w, h) \quad \forall l_i \in L$.

Finally, we combine these stickers into a single region-sticker mask $M_{\text{sticker}} = \bigcup_{i} B_i$, where every pixel inside the region has a larger perturbation budget than the rest of the mask:

\begin{equation}
    \epsilon_{sticker}(x, y, c) = \begin{cases}
\epsilon_A & \text{if } (x, y, c) \in M_{\text{sticker}} \\
\epsilon & \text{otherwise}
\end{cases}
\end{equation}

Region sticker focusing function $A_{sticker}$ thus applies variable perturbation budget $\epsilon_{sticker}(x, y, c)$ to every pixel with coordinates $(x, y)$ and channel $c$ in a perturbed synthetic image $S_n^k + \delta^k$ during optimization of loss $\mathcal{L}$ in equation \ref{eq:loss}.

\subsubsection{High-Pass Mask} We observe that noise applied to low-frequency regions of an image, such as a person's cheek, is much more visible than that applied to high-frequency regions. We hypothesize that hiding larger magnitudes of noise within high-frequency regions will enable stronger protection without sacrificing defense stealthiness. Thus, we propose our second focusing function $A_{highpass}$. First, we define a high-pass filter $H(I)$ normalized to mean zero and standard deviation one:

\begin{equation}
    H(I) = \frac{I - \mathcal{K}_{\sigma} * I}{\sigma(I - \mathcal{K}_\sigma * I)}
\end{equation}

Where $\mathcal{K}_{\sigma}$ is a Gaussian kernel with standard deviation $\sigma$, and $\sigma(\cdot)$ computes  standard deviation. Similar to Region-Stickers, we apply $A_{highpass}$ with a piecewise projection perturbation budget according to:

\begin{equation}
    \epsilon_{highpass}(x, y, c) = \begin{cases}
\epsilon_A & \text{if } H(I) > \mu \\
\epsilon & \text{otherwise}
\end{cases}
\end{equation}

Where $\mu$ is a threshold parameter. The high-pass mask gives only high frequency pixels above the threshold a larger perturbation budget, allowing the optimization to place more noise in those areas.

\begin{table*}[]
\centering
\caption{Identity-specific methods comparison. Protection Success Rate (PSR \%) of \textsc{FaceCloak} defense under the black-box face identification task on the Privacy-Commons dataset. Bolded and in blue is best, underlined is second best for each surrogate model and each target model. $\uparrow/\downarrow$ show the difference in performance between \textsc{FaceCloak} and the next best method.}
\label{tab:common-results}
\resizebox{\textwidth}{!}{%
\begin{tabular}{c|c|cc|cc|cc|cc|cc|cc|cc}

\toprule
\multirow{2}{*}{\textbf{Surrogate Model}} & \multirow{2}{*}{\textbf{Method}} & \multicolumn{2}{c|}{\textbf{ArcFace}} & \multicolumn{2}{c|}{\textbf{CosFace}} & \multicolumn{2}{c|}{\textbf{SFace}} & \multicolumn{2}{c|}{\textbf{MobileNet}} & \multicolumn{2}{c|}{\textbf{SENet}} & \multicolumn{2}{c|}{\textbf{IR50}} & \multicolumn{2}{c}{\textbf{Average}} \\
& & \textbf{Top-1} & \textbf{Top-5} & \textbf{Top-1} & \textbf{Top-5} & \textbf{Top-1} & \textbf{Top-5} & \textbf{Top-1} & \textbf{Top-5} & \textbf{Top-1} & \textbf{Top-5} & \textbf{Top-1} & \textbf{Top-5} & \textbf{Top-1} & \textbf{Top-5} \\

\midrule

\multirow{12}{*}{ArcFace}
& GD-UAP \cite{gduap-mopuri-2018} & 6.3 & 2.7 & 3.1 & 1.6 & 3.0 & 1.5 & 8.7 & 3.5 & 5.8 & 2.7 & 3.0 & 1.5 & 5.0 & 2.3 \\
& GAP \cite{poursaeed2018gap} & 30.7 & 20.7 & 19.7 & 13.9 & 23.9 & 15.9 & 33.6 & 20.8 & 31.0 & 20.1 & 13.1 & 7.4 & 25.3 & 16.5 \\
& AdvFaces+ \cite{Deb2019AdvFaces} & 75.1 & 65.7 & 69.3 & 60.7 & 73.0 & 64.5 & 77.0 & 65.9 & 74.5 & 63.4 & 71.2 & 60.8 & 73.4 & 63.5 \\
& FI-UAP \cite{zhong2022OPOM} & 72.3 & 62.4 & 63.5 & 53.3 & 70.3 & 61.9 & 73.9 & 61.8 & 77.4 & 67.6 & 52.4 & 40.7 & 68.3 & 58.0 \\
& OPOM-Affine Hull \cite{zhong2022OPOM} & 73.0 & 63.1 & 63.3 & 53.8 & 71.1 & 62.0 & 74.7 & 63.0 & 77.8 & 68.6 & 52.7 & 40.8 & 68.8 & 58.6 \\
& OPOM-Class Center \cite{zhong2022OPOM} & 76.9 & 67.8 & 69.2 & 60.3 & 75.0 & 67.3 & 78.4 & 67.9 & 82.1 & 73.4 & 57.1 & 45.6 & 73.1 & 63.7 \\
& OPOM-Convex Hull \cite{zhong2022OPOM} & 78.0 & 69.4 & 70.2 & 61.4 & 76.1 & 68.7 & 79.2 & 69.1 & \underline{82.9} & \underline{74.2} & 58.7 & 47.2 & 74.2 & 65.0 \\
& AdvCloak \cite{liu2023advcloak} & 80.1 & 71.0 & 74.7 & 65.7 & 78.0 & 69.7 & 82.0 & 71.9 & 78.1 & 67.4 & 77.4 & 67.5 & 78.4 & 68.9 \\
& AdvCloak-Affine Hull \cite{liu2023advcloak} & 79.5 & 70.6 & 74.1 & 65.1 & 77.5 & 69.2 & 81.9 & 71.7 & 77.9 & 67.4 & 77.2 & 67.2 & 78.0 & 68.5 \\
& AdvCloak-Class Center \cite{liu2023advcloak} & 81.1 & 72.3 & 75.8 & 67.5 & 79.0 & 71.1 & \cellcolor{blue!10}\textbf{83.8} & \underline{74.7} & 80.2 & 70.9 & 79.0 & 69.4 & 79.8 & \underline{71.0} \\
& AdvCloak-Convex Hull \cite{liu2023advcloak} & \underline{81.1} & \underline{72.7} & \underline{75.9} & \underline{67.5} & \underline{79.1} & \underline{71.3} & \underline{83.7} & 74.3 & 79.8 & 70.2 & \underline{79.0} & \underline{69.4} & \underline{79.8} & 70.9 \\
& Ours & \cellcolor{blue!10}\textbf{90.0}$_{\textcolor{green!50!black}{\uparrow8.9}}$ & \cellcolor{blue!10}\textbf{86.2}$_{\textcolor{green!50!black}{\uparrow13.5}}$ & \cellcolor{blue!10}\textbf{77.2}$_{\textcolor{green!50!black}{\uparrow1.3}}$ & \cellcolor{blue!10}\textbf{70.1}$_{\textcolor{green!50!black}{\uparrow2.6}}$ & \cellcolor{blue!10}\textbf{80.5}$_{\textcolor{green!50!black}{\uparrow1.4}}$ & \cellcolor{blue!10}\textbf{71.7}$_{\textcolor{green!50!black}{\uparrow0.4}}$ & 82.2$_{\textcolor{red!50!black}{\downarrow1.6}}$ & \cellcolor{blue!10}\textbf{76.0}$_{\textcolor{green!50!black}{\uparrow1.3}}$ & \cellcolor{blue!10}\textbf{85.9}$_{\textcolor{green!50!black}{\uparrow3.0}}$ & \cellcolor{blue!10}\textbf{77.3}$_{\textcolor{green!50!black}{\uparrow3.1}}$ & \cellcolor{blue!10}\textbf{96.9}$_{\textcolor{green!50!black}{\uparrow17.9}}$ & \cellcolor{blue!10}\textbf{94.7}$_{\textcolor{green!50!black}{\uparrow25.3}}$ & \cellcolor{blue!10}\textbf{85.4}$_{\textcolor{green!50!black}{\uparrow5.6}}$ & \cellcolor{blue!10}\textbf{79.3}$_{\textcolor{green!50!black}{\uparrow8.3}}$ \\
\midrule

\multirow{12}{*}{CosFace}
& GD-UAP \cite{gduap-mopuri-2018} & 3.7 & 1.4 & 1.1 & 0.4 & 1.3 & 0.4 & 4.0 & 1.5 & 3.2 & 1.0 & 1.8 & 0.6 & 2.5 & 0.9 \\
& GAP \cite{poursaeed2018gap} & 42.3 & 30.6 & 25.4 & 16.5 & 23.7 & 16.3 & 29.6 & 18.0 & 31.7 & 21.1 & 16.8 & 10.1 & 28.2 & 18.8 \\
& AdvFaces+ \cite{Deb2019AdvFaces} & 73.3 & 63.1 & 70 & 61.4 & 72.1 & 63.3 & 72.0 & 59.3 & 70.2 & 58.0 & 67.0 & 56.6 & 70.8 & 60.3 \\
& FI-UAP \cite{zhong2022OPOM} & 75.1 & 65.8 & 67.0 & 57.9 & 70.7 & 62.0 & 48.6 & 33.4 & 60.0 & 46.4 & 41.5 & 29.6 & 60.5 & 49.2 \\
& OPOM-Affine Hull \cite{zhong2022OPOM} & 76.1 & 67.4 & 68.8 & 60.5 & 72.5 & 63.9 & 50.2 & 35.2 & 61.4 & 49.3 & 43.7 & 31.8 & 62.1 & 51.3 \\
& OPOM-Class Center \cite{zhong2022OPOM} & 80.1 & 71.8 & 72.6 & 64.9 & 76.5 & 69.1 & 53.6 & 38.9 & 66.2 & 53.5 & 47.2 & 34.9 & 66.0 & 55.5 \\
& OPOM-Convex Hull \cite{zhong2022OPOM} & \underline{81.4} & \underline{73.2} & 73.9 & 66.0 & \underline{77.6} & \underline{70.7} & 54.8 & 39.6 & 67.3 & 54.7 & 48.4 & 36.3 & 67.2 & 56.7 \\
& AdvCloak \cite{liu2023advcloak} & 78.4 & 68.5 & 74.3 & 65.4 & 76.5 & 68.1 & 76.6 & 64.9 & 73.6 & 61.4 & 72.1 & 61.2 & 75.3 & 64.9 \\
& AdvCloak-Affine Hull \cite{liu2023advcloak} & 77.7 & 68.1 & 74.0 & 65.3 & 76.3 & 67.7 & 76.7 & 64.7 & 73.5 & 61.2 & 72.1 & 61.4 & 75.0 & 64.7 \\
& AdvCloak-Class Center \cite{liu2023advcloak} & 78.7 & 69.5 & 75.4 & \underline{67.4} & 77.5 & 69.6 & \underline{78.6} & \underline{68.0} & \underline{76.0} & \underline{64.8} & \underline{74.4} & \underline{64.3} & \underline{76.8} & \underline{67.3} \\
& AdvCloak-Convex Hull \cite{liu2023advcloak} & 79.1 & 69.7 & \underline{75.7} & 67.0 & 77.4 & 69.6 & 78.1 & 66.8 & 75.2 & 63.6 & 73.6 & 63.0 & 76.5 & 66.6 \\
& Ours & \cellcolor{blue!10}\textbf{88.8}$_{\textcolor{green!50!black}{\uparrow7.4}}$ & \cellcolor{blue!10}\textbf{84.5}$_{\textcolor{green!50!black}{\uparrow11.3}}$ & \cellcolor{blue!10}\textbf{84.1}$_{\textcolor{green!50!black}{\uparrow8.4}}$ & \cellcolor{blue!10}\textbf{77.4}$_{\textcolor{green!50!black}{\uparrow10.0}}$ & \cellcolor{blue!10}\textbf{81.3}$_{\textcolor{green!50!black}{\uparrow3.7}}$ & \cellcolor{blue!10}\textbf{73.0}$_{\textcolor{green!50!black}{\uparrow2.3}}$ & \cellcolor{blue!10}\textbf{82.2}$_{\textcolor{green!50!black}{\uparrow3.6}}$ & \cellcolor{blue!10}\textbf{75.7}$_{\textcolor{green!50!black}{\uparrow7.7}}$ & \cellcolor{blue!10}\textbf{86.2}$_{\textcolor{green!50!black}{\uparrow10.2}}$ & \cellcolor{blue!10}\textbf{77.5}$_{\textcolor{green!50!black}{\uparrow12.7}}$ & \cellcolor{blue!10}\textbf{97.5}$_{\textcolor{green!50!black}{\uparrow23.1}}$ & \cellcolor{blue!10}\textbf{95.5}$_{\textcolor{green!50!black}{\uparrow31.2}}$ & \cellcolor{blue!10}\textbf{86.7}$_{\textcolor{green!50!black}{\uparrow9.9}}$ & \cellcolor{blue!10}\textbf{80.6}$_{\textcolor{green!50!black}{\uparrow13.3}}$ \\
\midrule

\multirow{12}{*}{Softmax}
& GD-UAP \cite{gduap-mopuri-2018} & 6.3 & 2.7 & 3.1 & 1.6 & 3.0 & 1.5 & 8.7 & 3.5 & 5.8 & 2.7 & 0.3 & 1.5 & 4.5 & 2.3 \\
& GAP \cite{poursaeed2018gap} & 17.8 & 11.4 & 11.1 & 7.6 & 12.5 & 7.6 & 21.7 & 12.0 & 18.8 & 11.1 & 7 & 3.8 & 14.8 & 8.9 \\
& AdvFaces+ \cite{Deb2019AdvFaces} & 75.3 & 65.5 & 70.0 & 61.5 & 73.4 & 65.0 & 80.7 & 71.3 & 77 & 67.2 & 72.1 & 61.8 & 74.7 & 65.4 \\
& FI-UAP \cite{zhong2022OPOM} & 65.2 & 53.6 & 55.7 & 45.4 & 63.1 & 53.3 & 66.7 & 53.2 & 70.5 & 59.6 & 44.3 & 33.0 & 60.9 & 49.7 \\
& OPOM-Affine Hull \cite{zhong2022OPOM} & 65.9 & 54.6 & 56.1 & 46.0 & 63.8 & 53.4 & 67.8 & 54.5 & 71.7 & 60.3 & 44.0 & 32.8 & 61.5 & 50.3 \\
& OPOM-Class Center \cite{zhong2022OPOM} & 69.0 & 58.7 & 60.7 & 51.0 & 67.0 & 59.0 & 71.7 & 59.5 & 75.5 & 65.3 & 48.8 & 37.3 & 65.6 & 55.2 \\
& OPOM-Convex Hull \cite{zhong2022OPOM} & 70.7 & 60.2 & 61.8 & 52.1 & 69.4 & 60.5 & 72.6 & 60.7 & 76.2 & 66.4 & 50.2 & 38.7 & 66.8 & 56.4 \\
& AdvCloak \cite{liu2023advcloak} & 79.0 & 69.3 & 73.8 & 64.6 & 77.9 & 69.6 & 84.0 & 75.0 & 79.7 & 70.5 & 77.2 & 66.3 & 78.6 & 69.2 \\
& AdvCloak-Affine Hull \cite{liu2023advcloak} & 79.2 & 69.8 & 74.1 & 64.8 & 78.0 & 69.5 & 83.7 & 75.1 & 79.9 & 70.7 & 76.8 & 66.3 & 78.6 & 69.4 \\
& AdvCloak-Class Center \cite{liu2023advcloak} & \underline{81.6} & 73.0 & \underline{76.9} & \underline{68.6} & \underline{80.5} & \underline{72.3} & \cellcolor{blue!10}\textbf{87.3} & \cellcolor{blue!10}\textbf{79.3} & \underline{83.2} & \underline{75.6} & \underline{80.4} & \underline{70.9} & \underline{81.7} & \underline{73.3} \\
& AdvCloak-Convex Hull \cite{liu2023advcloak} & 80.7 & \underline{71.7} & 75.3 & 66.8 & 79.6 & 71.2 & \underline{86.1} & \underline{78.2} & 82.7 & 74.1 & 79.1 & 69.3 & 80.6 & 71.9 \\
& Ours & \cellcolor{blue!10}\textbf{90.0}$_{\textcolor{green!50!black}{\uparrow8.4}}$ & \cellcolor{blue!10}\textbf{86.4}$_{\textcolor{green!50!black}{\uparrow14.7}}$ & \cellcolor{blue!10}\textbf{80.7}$_{\textcolor{green!50!black}{\uparrow3.8}}$ & \cellcolor{blue!10}\textbf{73.8}$_{\textcolor{green!50!black}{\uparrow5.2}}$ & \cellcolor{blue!10}\textbf{90.3}$_{\textcolor{green!50!black}{\uparrow9.8}}$ & \cellcolor{blue!10}\textbf{85.0}$_{\textcolor{green!50!black}{\uparrow12.7}}$ & 82.4$_{\textcolor{red!40!black}{\downarrow4.9}}$ & 76.4$_{\textcolor{red!40!black}{\downarrow2.9}}$ & \cellcolor{blue!10}\textbf{88.3}$_{\textcolor{green!50!black}{\uparrow5.1}}$ & \cellcolor{blue!10}\textbf{81.0}$_{\textcolor{green!50!black}{\uparrow5.4}}$ & \cellcolor{blue!10}\textbf{97.9}$_{\textcolor{green!50!black}{\uparrow17.5}}$ & \cellcolor{blue!10}\textbf{96.7}$_{\textcolor{green!50!black}{\uparrow25.8}}$ & \cellcolor{blue!10}\textbf{88.3}$_{\textcolor{green!50!black}{\uparrow6.6}}$ & \cellcolor{blue!10}\textbf{83.2}$_{\textcolor{green!50!black}{\uparrow9.9}}$ \\

\bottomrule
\end{tabular}%
}
\end{table*}

\subsubsection{Learnable Attention}
Both $A_{sticker}$ and $A_{highpass}$ make assumptions about pixel importance prior to perturbation generation. We propose a third focusing function, $A_{attention}$, that learns which pixels are most important to face privacy protection as perturbation generation iterates. We hypothesize that these important pixels are non-obvious and may not be localized to key facial features or high frequency regions, meaning they should be learned instead. 

We initialize a per-pixel attention map $\alpha$ according to a uniform random distribution. As perturbation generation iterates, we update $\alpha$ according to:

\begin{equation}
    \alpha_{(t+1)} \leftarrow \alpha_t - z_{\alpha}\zeta(\frac{\partial\mathcal{L}(\mathcal{S}, G^+, G^-, \delta^k, A)}{\partial\alpha_t})
\end{equation}

Where $z_{\alpha}$ is the learning rate for the attention map and $\zeta(\cdot)$ is a normalization function. We show how to combine this learnable attention map with the perturbation in the next section.

\subsubsection{Combining Focal Masks}
Our Region-Sticker, High-Pass, and Learnable Attention masks can be applied individually or in combination. Region-Sticker and High-Pass both operate on the perturbation budget $\epsilon$ used by projection operator $\Pi(\cdot)$ during loss optimization. If these regions overlap on a pixel, then the perturbation budget for that pixel is $\epsilon_A$. Learnable attention is applied via element-wise multiplication with the perturbation before the projection operation. We apply these three focusing mechanisms to the perturbation and optimize the loss in Equation \ref{eq:loss} via iterative optimization similar to PGD \cite{madry2019pgd}. First, we update the perturbation according to:

\begin{equation}
\delta_{(t+1)}^k \leftarrow \delta^{k}_{t} - \lambda \cdot sign(\nabla_\delta \mathcal{L})
\end{equation}

Where $\lambda$ is the step size. Then, we calculate the updated perturbed synthetic image:

\begin{equation}
    S_{t+1}^k = S_{t}^k + \Pi (\delta_t^k \odot \alpha_t^k, \epsilon, -\epsilon)
\end{equation}

Where $\Pi$ is a projection function onto a perturbation budget. This hyperparameter is either the baseline value $\epsilon$ or a larger value $\epsilon_A$ according to the Region-Sticker and High-Pass Masks for each pixel with coordinates $(x, y)$ and channel $c$:

\begin{equation}
    \epsilon_{total}(x, y, c) = \epsilon_{sticker}(x, y, c) \cup \epsilon_{highpass}(x, y, c)
\end{equation}

We then repeat this process iteratively until the number of iterations is reached.

\subsection{Inference-Time Identity Protection}
Synthetic image generation and identity-specific perturbation optimization are offline procedures. Once $\delta^k$ is computed, it can be easily applied to any user face image $I_0^k$ via simple element-wise addition. A user only needs to submit a single image of themselves to the perturbation generation process once and they will obtain an identity-specific mask that can be applied without ever sending another image and with trivial computation. We conduct experiments to demonstrate that this mask is effective across a variety of facial recognition models, datasets, and face image examples.

\section{Experiments}
\subsection{Implementation Details}

\begin{table*}[htbp]
\centering
\caption{Identity-specific methods comparison. Protection Success Rate (PSR \%) of \textsc{FaceCloak} defense under the black-box face identification task on the Privacy-Celebs dataset. Bolded and in blue is best, underlined is second best for each surrogate model and each target model. $\uparrow/\downarrow$ show the difference in performance between \textsc{FaceCloak} and the next best method.}
\label{tab:celeb-results}
\resizebox{\textwidth}{!}{%
\begin{tabular}{c|c|cc|cc|cc|cc|cc|cc|cc}

\toprule
\multirow{2}{*}{\textbf{Surrogate Model}} & \multirow{2}{*}{\textbf{Method}} & \multicolumn{2}{c|}{\textbf{ArcFace}} & \multicolumn{2}{c|}{\textbf{CosFace}} & \multicolumn{2}{c|}{\textbf{SFace}} & \multicolumn{2}{c|}{\textbf{MobileNet}} & \multicolumn{2}{c|}{\textbf{SENet}} & \multicolumn{2}{c|}{\textbf{IR50}} & \multicolumn{2}{c}{\textbf{Average}} \\
& & \textbf{Top-1} & \textbf{Top-5} & \textbf{Top-1} & \textbf{Top-5} & \textbf{Top-1} & \textbf{Top-5} & \textbf{Top-1} & \textbf{Top-5} & \textbf{Top-1} & \textbf{Top-5} & \textbf{Top-1} & \textbf{Top-5} & \textbf{Top-1} & \textbf{Top-5} \\

\midrule

\multirow{12}{*}{ArcFace}
& GD-UAP \cite{gduap-mopuri-2018} & 6.5 & 2.4 & 3.4 & 1.2 & 4.0 & 1.3 & 10.0 & 3.8 & 6.4 & 2.5 & 3.8 & 1.3 & 5.7 & 2.1 \\
& GAP \cite{poursaeed2018gap} & 40.8 & 30.9 & 29.4 & 21.5 & 34.8 & 26.3 & 45.6 & 33.6 & 39.0 & 28.3 & 26.0 & 17.0 & 35.9 & 26.3 \\
& AdvFaces+ \cite{Deb2019AdvFaces} & 58.4 & 46.6 & 51.8 & 40.6 & 56.2 & 45.3 & 62.3 & 47.6 & 58.8 & 45.0 & 52.1 & 38.0 & 56.6 & 43.9 \\
& FI-UAP \cite{zhong2022OPOM} & 61.9 & 51.2 & 47.7 & 37.4 & 55.3 & 45.5 & 50.9 & 36.2 & 54.3 & 41.2 & 36.2 & 25.4 & 51.0 & 39.5 \\
& OPOM-Affine Hull \cite{zhong2022OPOM} & 64.9 & 54.6 & 50.6 & 40.8 & 58.1 & 48.9 & 53.3 & 38.5 & 57.2 & 44.4 & 38.6 & 27.3 & 53.8 & 42.4 \\
& OPOM-Class Center \cite{zhong2022OPOM} & 66.9 & 57.5 & 53.8 & 44.5 & 61.7 & 53.0 & 55.1 & 41.1 & 60.4 & 48.1 & 42.0 & 30.5 & 56.7 & 45.8 \\
& OPOM-Convex Hull \cite{zhong2022OPOM} & 68.5 & \underline{59.1} & 55.0 & 45.9 & 63.0 & 54.0 & 56.4 & 41.9 & 61.7 & 49.4 & 43.3 & 31.7 & 58.0 & 47.0 \\
& AdvCloak \cite{liu2023advcloak} & 67.7 & 56.9 & 60.4 & 50.3 & 64.6 & 55.1 & 71.8 & 60.0 & 67.2 & 55.0 & 62.4 & 50.3 & 65.7 & 54.6 \\
& AdvCloak-Affine Hull \cite{liu2023advcloak} & 66.8 & 55.9 & 59.8 & 49.2 & 63.6 & 53.9 & 70.3 & 57.8 & 65.8 & 53.2 & 61.1 & 48.6 & 64.6 & 53.1 \\
& AdvCloak-Class Center \cite{liu2023advcloak} & 67.6 & 56.9 & 61.0 & 50.9 & 64.9 & 55.3 & 72.3 & 60.0 & 67.9 & 55.8 & 62.5 & 50.4 & 66.0 & 54.9 \\
& AdvCloak-Convex Hull \cite{liu2023advcloak} & \underline{68.5} & 57.6 & \underline{61.4} & \underline{51.3} & \underline{65.3} & \underline{55.8} & \underline{72.3} & \underline{60.3} & \underline{67.9} & \underline{55.8} & \underline{62.9} & \underline{50.7} & \underline{66.4} & \underline{55.2} \\
& Ours & \cellcolor{blue!10}\textbf{82.2}$_{\textcolor{green!50!black}{\uparrow13.7}}$ & \cellcolor{blue!10}\textbf{76.3}$_{\textcolor{green!50!black}{\uparrow17.2}}$ & \cellcolor{blue!10}\textbf{72.7}$_{\textcolor{green!50!black}{\uparrow11.3}}$ & \cellcolor{blue!10}\textbf{65.6}$_{\textcolor{green!50!black}{\uparrow14.3}}$ & \cellcolor{blue!10}\textbf{78.9}$_{\textcolor{green!50!black}{\uparrow13.6}}$ & \cellcolor{blue!10}\textbf{73.2}$_{\textcolor{green!50!black}{\uparrow17.4}}$ & \cellcolor{blue!10}\textbf{86.7}$_{\textcolor{green!50!black}{\uparrow14.4}}$ & \cellcolor{blue!10}\textbf{83.6}$_{\textcolor{green!50!black}{\uparrow23.3}}$ & \cellcolor{blue!10}\textbf{79.5}$_{\textcolor{green!50!black}{\uparrow11.6}}$ & \cellcolor{blue!10}\textbf{72.5}$_{\textcolor{green!50!black}{\uparrow16.7}}$ & \cellcolor{blue!10}\textbf{85.1}$_{\textcolor{green!50!black}{\uparrow22.2}}$ & \cellcolor{blue!10}\textbf{80.6}$_{\textcolor{green!50!black}{\uparrow29.9}}$ & \cellcolor{blue!10}\textbf{80.9}$_{\textcolor{green!50!black}{\uparrow14.5}}$ & \cellcolor{blue!10}\textbf{75.3}$_{\textcolor{green!50!black}{\uparrow20.1}}$ \\

\midrule

\multirow{12}{*}{CosFace}
& GD-UAP \cite{zhang2025ctuap} & 7.4 & 3.0 & 3.7 & 1.3 & 4.2 & 1.4 & 10.2 & 4.2 & 6.3 & 2.4 & 3.8 & 1.3 & 5.9 & 2.3 \\
& GAP \cite{poursaeed2018gap} & 42.5 & 32.6 & 30.4 & 21.7 & 32.1 & 23.1 & 39.6 & 26.6 & 37.4 & 26.9 & 17.2 & 9.5 & 33.2 & 23.4 \\
& AdvFaces+ \cite{Deb2019AdvFaces} & 57.9 & 44.7 & 52.1 & 40.9 & 55.6 & 44.1 & 58.1 & 43.1 & 54.8 & 40.4 & 49.8 & 35.3 & 54.7 & 41.4 \\
& FI-UAP \cite{zhong2022OPOM} & 62.1 & 50.8 & 50.4 & 40.3 & 55.9 & 46.1 & 46.1 & 31.4 & 48.8 & 36.0 & 32.5 & 21.2 & 49.3 & 37.6 \\
& OPOM-Affine Hull \cite{zhong2022OPOM} & 65.0 & 54.3 & 54.2 & 43.9 & 59.7 & 49.9 & 49.1 & 34.2 & 52.6 & 39.5 & 34.8 & 23.4 & 52.6 & 40.9 \\
& OPOM-Class Center \cite{zhong2022OPOM} & 66.0 & 56.2 & 56.6 & 46.5 & 61.7 & 52.5 & 49.5 & 35.0 & 54.3 & 41.4 & 36.4 & 25.1 & 54.1 & 42.8 \\
& OPOM-Convex Hull \cite{zhong2022OPOM} & 67.4 & \underline{58.0} & 57.9 & 47.9 & 63.2 & 54.1 & 50.8 & 35.9 & 56.1 & 43.2 & 37.8 & 26.3 & 55.5 & 44.2 \\
& AdvCloak \cite{liu2023advcloak} & 67.3 & 56.5 & 60.8 & 50.2 & 64.4 & 54.1 & 69.2 & \underline{56.5} & 64.1 & 51.2 & 58.5 & 45.2 & 64.1 & 52.3 \\
& AdvCloak-Affine Hull \cite{liu2023advcloak} & 66.7 & 55.4 & 60.5 & 49.9 & 64.2 & 53.4 & 68.3 & 55.2 & 63.4 & 50.4 & 57.9 & 44.4 & 63.5 & 51.5 \\
& AdvCloak-Class Center \cite{liu2023advcloak} & 66.5 & 54.9 & 61.1 & 50.3 & 64.3 & 54.1 & 68.2 & 55.1 & 64.7 & 51.7 & 59.1 & 45.4 & 64.0 & 51.9 \\
& AdvCloak-Convex Hull \cite{liu2023advcloak} & \underline{67.4} & 57.6 & \underline{61.6} & \underline{51.0} & \underline{64.6} & \underline{54.6} & \underline{69.2} & 55.9 & \underline{65.0} & \underline{51.9} & \underline{59.1} & \underline{45.8} & \underline{64.5} & \underline{52.6} \\
& Ours & \cellcolor{blue!10}\textbf{78.8}$_{\textcolor{green!50!black}{\uparrow11.4}}$ & \cellcolor{blue!10}\textbf{73.1}$_{\textcolor{green!50!black}{\uparrow15.1}}$ & \cellcolor{blue!10}\textbf{75.5}$_{\textcolor{green!50!black}{\uparrow13.9}}$ & \cellcolor{blue!10}\textbf{68.8}$_{\textcolor{green!50!black}{\uparrow17.8}}$ & \cellcolor{blue!10}\textbf{79.1}$_{\textcolor{green!50!black}{\uparrow14.5}}$ & \cellcolor{blue!10}\textbf{73.3}$_{\textcolor{green!50!black}{\uparrow18.7}}$ & \cellcolor{blue!10}\textbf{86.8}$_{\textcolor{green!50!black}{\uparrow17.6}}$ & \cellcolor{blue!10}\textbf{83.6}$_{\textcolor{green!50!black}{\uparrow27.1}}$ & \cellcolor{blue!10}\textbf{79.6}$_{\textcolor{green!50!black}{\uparrow14.6}}$ & \cellcolor{blue!10}\textbf{72.7}$_{\textcolor{green!50!black}{\uparrow20.8}}$ & \cellcolor{blue!10}\textbf{85.6}$_{\textcolor{green!50!black}{\uparrow26.5}}$ & \cellcolor{blue!10}\textbf{81.1}$_{\textcolor{green!50!black}{\uparrow35.3}}$ & \cellcolor{blue!10}\textbf{80.9}$_{\textcolor{green!50!black}{\uparrow16.4}}$ & \cellcolor{blue!10}\textbf{75.4}$_{\textcolor{green!50!black}{\uparrow22.8}}$ \\

\midrule

\multirow{12}{*}{Softmax}
& GD-UAP \cite{zhang2025ctuap} & 6.5 & 2.4 & 3.4 & 1.2 & 3.6 & 1.3 & 10.7 & 3.8 & 5.7 & 2.1 & 3.1 & 0.9 & 5.5 & 2.0 \\
& GAP \cite{poursaeed2018gap} & 30.3 & 21.0 & 21.0 & 13.3 & 24.1 & 16.3 & 43.4 & 31.2 & 33.2 & 23.0 & 12.8 & 6.6 & 27.5 & 18.6 \\
& AdvFaces+ \cite{Deb2019AdvFaces} & 60.1 & 48.2 & 53.5 & 42.4 & 58.1 & 47.2 & 68.5 & 55.7 & 63.4 & 50.4 & 53.8 & 41.0 & 59.6 & 47.5 \\
& FI-UAP \cite{zhong2022OPOM} & 51.6 & 40.6 & 40.4 & 30.6 & 47.9 & 38.1 & 55.6 & 42.2 & 56.1 & 44.4 & 33.3 & 22.8 & 47.5 & 36.5 \\
& OPOM-Affine Hull \cite{zhong2022OPOM} & 53.3 & 42.7 & 42.6 & 32.6 & 49.7 & 40.1 & 58.0 & 44.5 & 58.5 & 46.4 & 35.1 & 23.8 & 49.5 & 38.4 \\
& OPOM-Class Center \cite{zhong2022OPOM} & 56.3 & 45.7 & 46.3 & 36.7 & 53.9 & 43.7 & 61.9 & 49.2 & 62.1 & 51.1 & 37.4 & 26.1 & 53.0 & 42.1 \\
& OPOM-Convex Hull \cite{zhong2022OPOM} & 58.3 & 47.8 & 48.0 & 38.4 & 55.6 & 45.6 & 62.8 & 50.3 & 63.7 & 52.5 & 38.9 & 27.7 & 54.6 & 43.7 \\
& AdvCloak \cite{liu2023advcloak} & 66.1 & 54.4 & 59.7 & 49.2 & 63.7 & 53.5 & 72.3 & 60.0 & 67.4 & 55.3 & 60.0 & 47.2 & 64.9 & 53.3 \\
& AdvCloak-Affine Hull \cite{liu2023advcloak} & 65.4 & 54.0 & 59.4 & 48.8 & 63.0 & 53.0 & 71.7 & 59.3 & 66.6 & 53.8 & 59.2 & 45.9 & 64.2 & 52.5 \\
& AdvCloak-Class Center \cite{liu2023advcloak} & \underline{68.3} & \underline{57.6} & \underline{62.1} & \underline{51.8} & \underline{66.1} & \underline{56.4} & \underline{75.9} & \underline{65.0} & \underline{71.5} & \underline{60.0} & \underline{63.0} & \underline{50.5} & \underline{67.8} & \underline{56.9} \\
& AdvCloak-Convex Hull \cite{liu2023advcloak} & 67.4 & 56.2 & 61.1 & 50.2 & 64.7 & 54.8 & 74.4 & 62.8 & 69.5 & 57.3 & 60.9 & 48.1 & 66.3 & 54.9 \\
& Ours & \cellcolor{blue!10}\textbf{78.7}$_{\textcolor{green!50!black}{\uparrow10.4}}$ & \cellcolor{blue!10}\textbf{73.1}$_{\textcolor{green!50!black}{\uparrow15.5}}$ & \cellcolor{blue!10}\textbf{73.0}$_{\textcolor{green!50!black}{\uparrow10.9}}$ & \cellcolor{blue!10}\textbf{65.6}$_{\textcolor{green!50!black}{\uparrow13.8}}$ & \cellcolor{blue!10}\textbf{81.6}$_{\textcolor{green!50!black}{\uparrow15.5}}$ & \cellcolor{blue!10}\textbf{76.5}$_{\textcolor{green!50!black}{\uparrow20.1}}$ & \cellcolor{blue!10}\textbf{86.9}$_{\textcolor{green!50!black}{\uparrow11.0}}$ & \cellcolor{blue!10}\textbf{83.4}$_{\textcolor{green!50!black}{\uparrow18.4}}$ & \cellcolor{blue!10}\textbf{80.1}$_{\textcolor{green!50!black}{\uparrow8.6}}$ & \cellcolor{blue!10}\textbf{74.1}$_{\textcolor{green!50!black}{\uparrow14.1}}$ & \cellcolor{blue!10}\textbf{85.9}$_{\textcolor{green!50!black}{\uparrow22.9}}$ & \cellcolor{blue!10}\textbf{82.3}$_{\textcolor{green!50!black}{\uparrow31.8}}$ & \cellcolor{blue!10}\textbf{81.1}$_{\textcolor{green!50!black}{\uparrow13.3}}$ & \cellcolor{blue!10}\textbf{75.8}$_{\textcolor{green!50!black}{\uparrow18.9}}$ \\

\bottomrule
\end{tabular}%
}
\end{table*}

\subsubsection{Evaluation Metrics} We measure the success of \textsc{FaceCloak} at protecting facial images on the 1:N identification task and on the 1:1 verification task. Both tasks use two datasets: a probe set and a gallery set. During evaluation, we select an image from the probe set. All other images of the same identity are temporarily added to the gallery set. The selected probe set image is then embedded with the face feature extraction model being evaluated and the probe embedding is compared to the embedding of each image in the gallery set. For Top-\textit{n} identification, the protection is successful if there is no image of the same identity as the probe image in the \textit{n} most similar images in the gallery set. For verification, the protection is successful if the face image identity is successfully changed to a different identity. We calculate Protection Success Rate (PSR) as the total number of images protected this way divided by the number of images in the probe set. We measure both Top-1 PSR and Top-5 PSR. 

\subsubsection{Datasets} We evaluate \textsc{FaceCloak} against other identity-specific methods on two face identification datasets: Privacy-Commons and Privacy-Celebrities \cite{zhong2022OPOM}. Privacy-Commons is comprised of 500 identities each with 20 images pulled from the MegaFace Challenge 2 dataset \cite{nech2017megaface2}. Of these, 5 images per identity form the probe set. Additionally, the Privacy-Commons gallery contains 10,000 other random MegaFace Challenge 2 images with no identity overlap to the probe set as distractors. To form the Privacy-Celebrities dataset we select 500 identities from MS-Celeb-1M \cite{guo2016msceleb1m} and add five distinct images per identity to the probe set. We add 13,233 images from the Labeled Faces in the Wild dataset \cite{huang2008lfw} to the gallery as distractors.

We evaluate \textsc{FaceCloak} against image-specific methods on face verification with the CelebA-HQ dataset \cite{liu2015celebahq}. We randomly select 1,000 identities with one unique image each and group these into 5 groups. These images are added to both the probe and gallery set. Our defense attempts to change each image in each group to a randomly selected identity, per the experimental settings of previous works \cite{Li2024GIFT, Wang2025AdvCpg, Sun2024diffam, shamshad2023clip2protect, hu2022amtgan}.

\subsubsection{Target FR Models} We test \textsc{FaceCloak} on ten different facial recognition models: ArcFace \cite{deng2018arcface}, CosFace \cite{wang2018cosface}, SFace \cite{boutros2022sface}, SENet \cite{hu2019senet}, MobileNet \cite{howard2017mobilenet}, and Inception Resnet-50 (IR50) for identity-specific comparisons, \cite{szegedy2016incresnet}, and Inception Resnet-152 (IR152) \cite{szegedy2016incresnet}, Inception Resnet SE 50 (IRSE50) \cite{szegedy2016incresnet}, Facenet \cite{schroff2015facenet} and MobileFace \cite{chen2018mobilefacenet} for image-specific comparisons.

\subsubsection{System Settings} For comparison with other identity-specific methods we select a perturbation budget $\epsilon = 8/255$ to allow for fair comparison with peer works \cite{zhong2022OPOM, Deb2019AdvFaces, liu2023advcloak, chow2024chameleon}. For image-specific comparisons we select a perturbation budget of $\epsilon = 12/255$ to match peer perturbation levels \cite{Wang2025AdvCpg, Sun2024diffam, hu2022amtgan}. We set the Region-Sticker and High-Pass Mask perturbation budget $\epsilon_A = 32 / 255$ for pixels selected by those methods. For each identity we generate a small set of eight synthetic faces. The universal perturbation is generated on each of these images over ten iterations with a step size of $2/255$.

\subsection{Identity-Specific Comparison Results} We compare \textsc{FaceCloak} with 11 representative identity-specific methods on Privacy-Commons in Table~\ref{tab:common-results}  and Privacy-Celebrities dataset in Table~\ref{tab:celeb-results}. ArcFace \cite{deng2018arcface}, CosFace \cite{wang2018cosface}, and Softmax are used for surrogate models. For each of these, protected images are evaluated against all six recognition models. Results from the 11 methods are cited from their original papers. \textsc{FaceCloak} outperforms all comparison methods on all models for both datasets with one exception on the Privacy-Commons dataset (Table~\ref{tab:common-results}): it shows decreased performance against the AdvCloak-ClassCenter \cite{liu2023advcloak} method with MobileNet model~\cite{howard2017mobilenet} under ArcFace and Softmax as the surrogate models. \textsc{FaceCloak} improves average Top-1 PSR by up to 9.9\% and average Top-5 PSR by up to 13.3\% on Privacy-Common, and it improves average Top-1 PSR by up to 16.4\% and average Top-5 PSR by up to 22.8\% on Privacy-Celeb. We make three additional observations. (1) \textsc{FaceCloak} defensive perturbations are highly transferable, showing strong protection success rates against all comparison facial recognition models regardless of which surrogate model is used. (2) Our method sees the most improvement on IR50 \cite{szegedy2016incresnet}, including 35.8\% improvement in Top-5 PSR on Privacy-Celebrities and a Top-1 PSR of 97.9\% from perturbations generated against Softmax on Privacy-Commons. (3) On average \textsc{FaceCloak} improves Top-5 PSR more than it improves Top-1 PSR on both datasets. This can be  beneficial in realistic scenarios, where an attacker may compare a user's face to multiple candidate faces instead of just the single most similar one.

\begin{figure}[h]
\centering
\includegraphics[width=0.8\columnwidth]{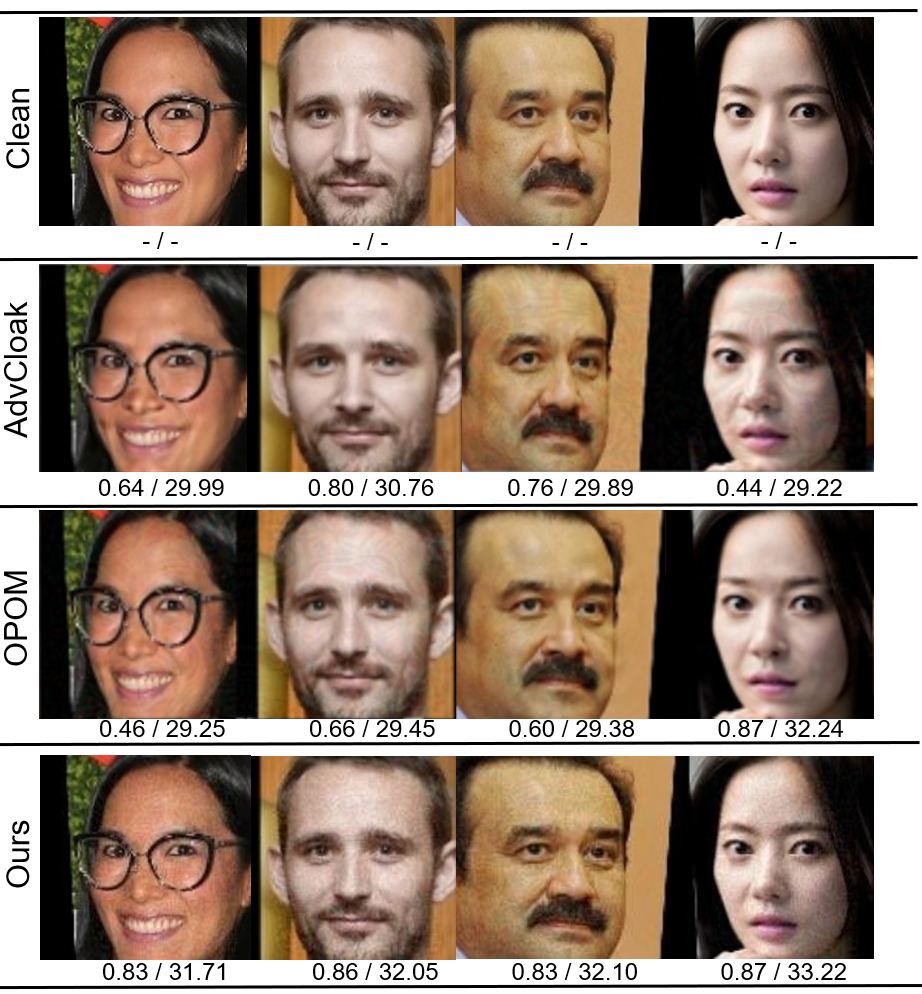}
\caption{Visual quality comparison with state of the art identity-specific protection methods. Values shown are SSIM ($\uparrow$) and PSNR ($\uparrow$) for that image with respect to the clean version. Images selected by \cite{liu2023advcloak}.}
\label{fig:universal-compare}
\end{figure}

\subsection{Image-Specific Comparison Results}
We compare \textsc{FaceCloak} with 18 representative image-specific methods on the face verification task on CelebA-HQ \cite{liu2015celebahq} in Table \ref{tab:image-specific-results}. We adapt our method to be identity-specific by generating a new perturbation for each image. Peer results are from \cite{Wang2025AdvCpg}. On IRSE50 \cite{szegedy2016incresnet}, Facenet \cite{schroff2015facenet}, and MobileFace \cite{chen2018mobilefacenet}, \textsc{FaceCloak} outperforms all other methods. Interestingly, \textsc{FaceCloak} is less effective than Adv-CPG \cite{Wang2025AdvCpg} on IR152. Overall, our method improves upon the state of the art by 3.6\% on average. 

\begin{table}[h]
\centering
\caption{Image-specific methods comparison. Top-1 Protection Success Rate (PSR \%) of \textsc{FaceCloak} defense under the face verification task on the CelebA-HQ dataset. Bolded in blue is best, underlined is second best for each model. $\uparrow/\downarrow$ show the difference between \textsc{FaceCloak} and the next best method.}
\label{tab:image-specific-results}
\resizebox{0.46\textwidth}{!}{%
\begin{tabular}{c|cccc|c}

\toprule
\textbf{Method} & \textbf{IR152} & \textbf{IRSE50} & \textbf{Facenet} & \textbf{MobileFace} & \textbf{Average} \\

\midrule
FGSM \cite{Goodfellow2014fgsm} & 12.1 & 45.8 & 1.4 & 53.0 & 28.1\\
MI-FGSM \cite{Dong2017mifgsm} & 46.5 & 70.6 & 27.1 & 58.9 & 50.8\\
PGD \cite{madry2019pgd} & 41.9 & 63.2 & 19.6 & 57.3 & 45.5\\
TI-DIM \cite{Dong2019tidim} & 35.1 & 62.4 & 13.7 & 52.8 & 41.3 \\
TIP-IM \cite{Yang2020tipim} & 41.3 & 57.3 & 39.1 & 49.6 & 46.8\\
Adv-Hat \cite{Komkov2019AdvHat} & 5.0 & 16.9 & 4.9 & 12.6 & 9.9 \\
Adv-Makeup \cite{yin2021advmakeup} & 12.7 & 20.0 & 1.4 & 22.1 & 14.0\\
AMT-GAN \cite{hu2022amtgan} & 12.1 & 53.3 & 4.9 & 48.0 & 29.5 \\
Clip2Protect \cite{shamshad2023clip2protect} & 47.6 & 81.0 & 42.6 & 73.6 & 61.2\\
GIFT \cite{Li2024GIFT} & 73.8 & 83.7 & 56.5 & 86.4 & 75.1 \\
DFPP \cite{shamsahd2025DFPP} & 46.4 & 80.6 & 45.4 & 72.1 & 61.1\\
DiffAM \cite{Sun2024diffam} & 65.1 & \underline{89.7} & 63.0 & 84.5 & 75.6\\
DiffProtect \cite{Liu2023DiffProtect} & 58.6 & 79.3 & 24.7 & 75.9 & 59.6\\
DPG \cite{zhang2024DPG} & 42.9 & 62.5 & 35.8 & 66.4 & 51.9\\
SD4Privacy \cite{An2024SD4Privacy} & 66.9 & 80.0 & 53.5 & 74.6 & 68.7\\
Adv-Diffusion \cite{liu2024advdiffusion} & 52.8 & 81.7 & 35.0 & 70.8 & 60.1\\
P3-Mask \cite{chow2024chameleon} & 73.5 & 83.4 & 60.2 & 69.6 & 71.7\\
Adv-CPG \cite{Wang2025AdvCpg} & \cellcolor{blue!10}\textbf{77.0} & 88.7 & \underline{63.5} & \underline{88.0} & \underline{79.3}\\
\midrule
Ours & $\underline{73.5}_{\textcolor{red!50!black}{\downarrow3.5}}$ & $\cellcolor{blue!10}\textbf{95.0}_{\textcolor{green!50!black}{\uparrow5.3}}$ & $\cellcolor{blue!10}\textbf{65.5}_{\textcolor{green!50!black}{\uparrow2.0}}$ & $\cellcolor{blue!10}\textbf{97.5}_{\textcolor{green!50!black}{\uparrow9.5}}$ & $\cellcolor{blue!10}\textbf{82.9}_{\textcolor{green!50!black}{\uparrow3.6}}$ \\

\bottomrule
\end{tabular}%
}
\end{table}

\begin{figure}[h]
\centering
\includegraphics[width=0.8\columnwidth]{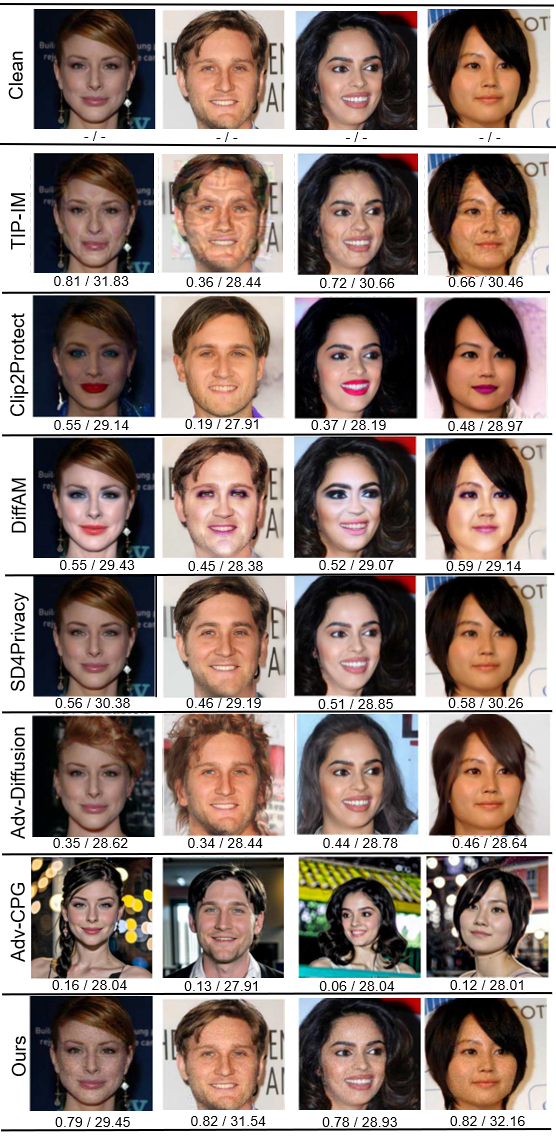}
\caption{Visual quality comparison with state of the art image-specific protection methods. Values shown are SSIM ($\uparrow$) and PSNR ($\uparrow$) for that image with respect to the clean version. Images selected by \cite{Wang2025AdvCpg}.}
\label{fig:single-compare}
\end{figure}

\section{Discussion}
\subsection{Visual Quality Comparison}
Perceptual quality is a key challenge for privacy-protecting perturbations. We show that, in addition to providing a strong defense, \textsc{FaceCloak} has stealthiness comparable to other identity-specific and image-specific methods. In Figure \ref{fig:universal-compare} we compare images protected by \textsc{FaceCloak} with images protected by OPOM \cite{zhong2022OPOM} and AdvCloak \cite{liu2023advcloak}. We also provide SSIM and PSNR perceptual quality metrics beneath each image to measure the perturbation's stealthiness. We compare \textsc{FaceCloak}'s perturbations to examples from state of the art image-specific methods in Figure \ref{fig:single-compare}. Although \textsc{FaceCloak} adds some noise to the images, it preserves the visual quality of the faces without altering their semantics. Finally, we statistically compare the perceptual quality of images protected by \textsc{FaceCloak} against seven image-specific methods on 1000 CelebA-HQ images in Table \ref{tab:percep-compare}. We use classical perceptual quality metrics SSIM and PSNR as well as deep neural network based metrics LPIPS \cite{zhang2018lpips}, DISTS \cite{ding2020dists}, and DreamSim \cite{fu2023dreamsim}. We note that \textsc{FaceCloak} achieves comparable perceptual quality to its peers while offering stronger protection (Table \ref{tab:image-specific-results}).



\begin{table}[]
\centering
\caption{Perceptual quality comparison against single-image methods on the CelebA-HQ dataset. Bolded in blue is best. Peer results from \cite{Wang2025AdvCpg}.}
\label{tab:percep-compare}
\resizebox{1.0\columnwidth}{!}{%

\begin{tabular}{c|ccccc}
\toprule
\textbf{Method} & \textbf{SSIM ($\uparrow$)} & \textbf{PSNR ($\uparrow$)} & \textbf{LPIPS ($\downarrow$)} & \textbf{DISTS ($\downarrow$)} & \textbf{DreamSim ($\downarrow$)} \\
\midrule
Clip2Protect \cite{shamshad2023clip2protect} & 0.60 & 19.33 & 0.30 & 0.186 & 0.31 \\
SD4Privacy \cite{An2024SD4Privacy} & 0.81 & 27.01 & \cellcolor{blue!10}\textbf{0.12} & \cellcolor{blue!10}\textbf{0.065} & 0.22 \\
TIP-IM \cite{Yang2020tipim} & \cellcolor{blue!10}\textbf{0.92} & \cellcolor{blue!10}\textbf{33.19} & 0.18 & 0.113 & 0.36 \\
AdvDiff \cite{liu2024advdiffusion} & 0.78 & 28.31 & 0.28 & 0.168 & \cellcolor{blue!10}\textbf{0.13} \\
DiffAM \cite{Sun2024diffam} & 0.88 & 20.12 & 0.14 & 0.083 & 0.41 \\
Adv-CPG \cite{Wang2025AdvCpg} & 0.86 & 29.88 & 0.44 & 0.237 & 0.48 \\
Ours & 0.81 & 32.63 & 0.13 & 0.083 & 0.21 \\
\bottomrule
\end{tabular}%
}
\end{table}

\subsection{Ablation Studies}
In Tables \ref{tab:abl1}-\ref{tab:abl3} we ablate the values of three key hyperparameters for our method: perturbation budget, number of iterations, and number of synthetic images. As expected, Table \ref{tab:abl1} shows that increasing perturbation budget increases protection success rate across all models. For our experiments, we choose perturbation budget values that match those used by peer studies to provide a fair comparison. In Table \ref{tab:abl2} we observe that increasing the number of iterations also increases average protection success rate up until around 10 iterations, after which performance remains roughly the same. For this reason, we select 10 as the number of iterations for our study. Similarly, the results in Table \ref{tab:abl3} show that 8 or more synthetic images have similar protection performance, so we select 8 as the value for our experiments.

\begin{table}[]
    \centering
    \caption{Ablation study of per-model protection success rate versus perturbation budget value for identity-specific methods.}
    \resizebox{\columnwidth}{!}{%
    \begin{tabular}{cccccccc}
         \toprule
         \textbf{Pert. Budget} & \textbf{ArcFace} & \textbf{CosFace} & \textbf{SFace} & \textbf{MobileNet} & \textbf{SENet} & \textbf{IR50} & \textbf{Avg.} \\
         \midrule
         2/255  & 74.8 & 61.3 & 60.9 & 71.8 & 58.0 & 75.2 & 67.0 \\
        4/255  & 77.3 & 61.7 & 61.9 & 71.9 & 64.0 & 82.4 & 69.9 \\
        6/255  & 79.5 & 64.1 & 62.2 & 71.5 & 69.6 & 88.4 & 72.3 \\
        8/255  & 81.1 & 63.8 & 64.5 & 72.0 & 77.8 & 90.6 & 75.0 \\
        10/255 & 81.8 & 67.5 & 67.8 & 71.8 & 81.8 & 95.5 & 77.7 \\
        12/255 & 87.5 & 70.4 & 70.5 & 73.3 & 88.5 & 99.3 & 81.6 \\
        14/255 & 88.4 & 71.2 & 71.6 & 73.4 & 90.2 & 97.8 & 82.1 \\
        16/255 & 94.3 & 76.6 & 75.9 & 73.6 & 94.4 & 98.9 & 85.6 \\
         \bottomrule
    \end{tabular}
    }
    \label{tab:abl1}
\end{table}

\begin{table}[]
    \centering
    \caption{Ablation study of per-model protection success rate versus number of iterations for identity-specific methods.}
    \resizebox{\columnwidth}{!}{%
    \begin{tabular}{cccccccc}
         \toprule
         \textbf{Iters.} & \textbf{ArcFace} & \textbf{CosFace} & \textbf{SFace} & \textbf{MobileNet} & \textbf{SENet} & \textbf{IR50} & \textbf{Avg.} \\
         \midrule
         2  & 44.7 & 27.3 & 27.8 & 39.4 & 35.0 & 53.2 & 37.9 \\
        4  & 66.1 & 47.8 & 45.1 & 54.8 & 57.8 & 74.1 & 57.6 \\
        6  & 71.6 & 54.2 & 54.5 & 64.0 & 66.2 & 82.0 & 65.4 \\
        8  & 76.6 & 57.0 & 55.8 & 65.5 & 69.4 & 85.2 & 68.3 \\
        10 & 81.1 & 63.8 & 64.4 & 72.0 & 77.8 & 90.6 & 75.0 \\
        12 & 83.2 & 68.1 & 62.3 & 72.2 & 77.2 & 94.3 & 76.2 \\
        14 & 81.7 & 64.9 & 61.4 & 75.5 & 78.2 & 93.3 & 75.8 \\
        16 & 80.5 & 63.6 & 63.6 & 72.1 & 74.6 & 91.8 & 74.3 \\
         \bottomrule
    \end{tabular}
    }
    \label{tab:abl2}
\end{table}

\begin{table}[]
    \centering
    \caption{Ablation study of per-model protection success rate versus number of synthetic images for identity-specific methods.}
    \resizebox{\columnwidth}{!}{%
    \begin{tabular}{cccccccc}
         \toprule
         \textbf{Num. Images} & \textbf{ArcFace} & \textbf{CosFace} & \textbf{SFace} & \textbf{MobileNet} & \textbf{SENet} & \textbf{IR50} & \textbf{Avg.} \\
         \midrule
         2  & 78.4 & 58.7 & 58.0 & 66.2 & 66.9 & 84.4 & 68.8 \\
        4  & 78.7 & 60.2 & 59.8 & 67.4 & 70.0 & 90.1 & 71.0 \\
        8  & 81.1 & 63.8 & 64.5 & 72.0 & 77.8 & 90.6 & 75.0 \\
        16 & 82.7 & 64.8 & 65.6 & 71.6 & 80.6 & 94.1 & 76.6 \\
        32 & 78.5 & 65.5 & 66.2 & 71.6 & 77.8 & 93.9 & 75.6 \\
        64 & 76.7 & 65.5 & 66.1 & 71.7 & 78.8 & 96.2 & 75.8 \\
         \bottomrule
    \end{tabular}
    }
    \label{tab:abl3}
\end{table}

We report additional ablation results on the different components of our method in Table \ref{tab:ablation}, showing the contributions of the individual components of \textsc{FaceCloak}. Baseline \textsc{FaceCloak} achieves performance comparable to other universal methods \cite{liu2023advcloak, zhong2022OPOM, Deb2019AdvFaces}. However, we observe that Region-Stickers (R), Learnable Attention (A), and High-Pass Masks (H) all contribute to improving the overall performance, with Region-Stickers achieving the largest improvement, followed by Learnable Attention and High-Pass Masks.  

\begin{table}[]
\centering
\caption{Ablation study comparing different components of \textsc{FaceCloak}. Top-1 Average and Top-5 Average show the average Protection Success Rate against six models on the indicated dataset. R indicates Region-Sticker, A indicates Attention, H indicates High-Pass Mask.}
\label{tab:ablation}
\resizebox{\columnwidth}{!}{%
\begin{tabular}{c|cc|cc}

\toprule
\multirow{2}{*}{\textbf{Method}} & \multicolumn{2}{c|}{\textbf{Privacy-Common}} & \multicolumn{2}{c}{\textbf{Privacy-Celebrities}} \\
 & \textbf{Top-1 Avg.} & \textbf{Top-5 Avg.} & \textbf{Top-1 Avg.} & \textbf{Top-5 Avg.} \\
\midrule
Baseline & 76.6 & 69.2 & 81.0 & 75.5 \\
Baseline + R & $85.8_{\textcolor{green!50!black}{\uparrow9.2}}$ & $79.5_{\textcolor{green!50!black}{\uparrow10.3}}$ & $86.7_{\textcolor{green!50!black}{\uparrow5.7}}$ & $82.3_{\textcolor{green!50!black}{\uparrow6.8}}$ \\
Baseline + R + A & $86.6_{\textcolor{green!50!black}{\uparrow0.8}}$ & $80.8_{\textcolor{green!50!black}{\uparrow1.3}}$ & $87.3_{\textcolor{green!50!black}{\uparrow0.6}}$ & $83.1_{\textcolor{green!50!black}{\uparrow0.8}}$ \\
\textsc{FaceCloak} (Baseline + R + A + H) & $86.8_{\textcolor{green!50!black}{\uparrow0.2}}$ & $81.0_{\textcolor{green!50!black}{\uparrow0.2}}$ & $87.5_{\textcolor{green!50!black}{\uparrow0.2}}$ & $83.3_{\textcolor{green!50!black}{\uparrow0.2}}$ \\

\bottomrule
\end{tabular}%
}
\end{table}

\begin{figure}
    \centering
    \includegraphics[width=\linewidth]{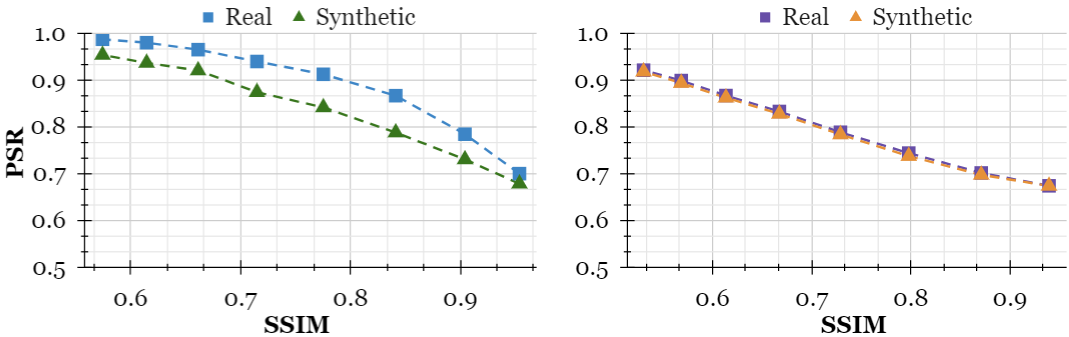}
    \caption{Perceptual quality and protection success rate tradeoff for eight real or eight synthetic images averaged over six models on Privacy-Commons. Perceptual quality varied by modulating perturbation budget $\epsilon$. Left: perturbations evaluated on same model as training. Right: perturbations transferred from MobileNet to five other models.}
    \label{fig:real-vs-synth}
\end{figure}

Next, we report our study on the differences between real and synthetic images in Figure \ref{fig:real-vs-synth}. We record the tradeoff between perceptual quality (SSIM) and protection effectiveness (PSR) averaged across six models on Privacy-Commons for eight real or eight synthetic images. We observe that when \textsc{FaceCloak} protective perturbations are evaluated on the same model they were trained on (left), real images perform slightly better. When the cloaks are transferred to other models (right), synthetic images achieve identical performance to real ones. This suggests that \textsc{FaceCloak}'s synthetic perturbation approach can achieve comparable protection to real image approaches without requiring multiple images from the user. We also note that even at very high perceptual quality levels (e.g. SSIM of 0.95) transferred cloaks achieve PSR scores of close to 70\%.

\subsection{Robustness to Adversarial Post-Processing}
\begin{figure}
    \centering
    \includegraphics[width=\linewidth]{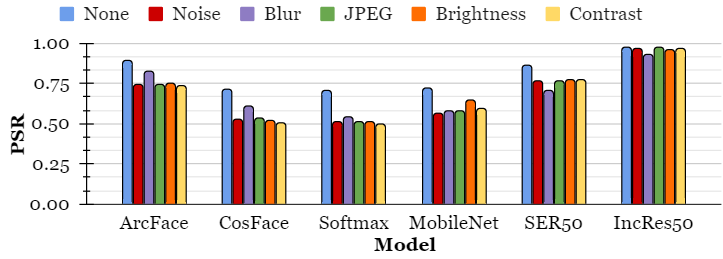}
    \caption{Robustness of \textsc{FaceCloak} perturbations under common image transformations. Performance calculated on Privacy-Commons images with an ArcFace surrogate model.}
    \label{fig:robustness}
\end{figure}

We investigate the performance of \textsc{FaceCloak}'s perturbations under common post-processing transformations including added noise, Gaussian blur, JPEG compression, brightness shift, and contrast shift in Figure \ref{fig:robustness}. We observe that perturbations optimized against ArcFace as a surrogate and evaluated on other models maintain their robustness under common image transformations. \textsc{FaceCloak}'s robustness to post-processing varies by transfer model and transformation. For example, perturbations show greater relative resistance to blurring for CosFace than SER50. 

\section{Conclusion}
We present \textsc{FaceCloak}, a three-stage system for learning facial privacy protection perturbations from a single user image. To our knowledge, ours is the first work to introduce synthetic image generation for identity-specific perturbations, achieving strong per-user protection without requiring multiple user images. Our ablation studies show that our novel face perturbation focusing mechanisms based on key facial landmarks and high frequency face regions increase protection. Experiments also show robustness to common post-processing transformations. Our identity-specific perturbations only need to be generated once for each user, after which they can easily be applied to any image for that user. Extensive experiments show that \textsc{FaceCloak} improves upon the state of the art in identity-specific and image-specific privacy protection perturbations while maintaining visual quality.

\section*{Impact Statement}
\textsc{FaceCloak} has the potential to support the digital privacy of individuals who wish to post pictures of themselves online. This is particularly relevant in light of the ever-growing popularity of social media, the widespread scraping of public data for training machine learning systems, and the general commodification of user data. By allowing users to protect all images of themselves with a single image, \textsc{FaceCloak} lowers the barrier to defending against unwanted facial recognition. While this work is not a catch-all solution to this complex issue, it is a step towards giving users control over their digital identities.

\section*{Acknowledgement}
This research is partially sponsored by the NSF CISE grants 2302720, 2312758, 2038029, an IBM faculty award, and a grant from CISCO Edge AI program. The first author is sponsored by Georgia Tech Research Institute PhD Fellowship Program. \textsc{FaceCloak} is open sourced at https://github.com/zacharyyahn/FaceCloak.

\bibliography{main}
\bibliographystyle{IEEEtran}

\newpage
\section{Biography Section}



{\bf Zachary Yahn} graduated from University of Virginia with a BS in Computer Science and in Computer Engineering and from Univerity College Dublin with a MS in Computer Science. He joined the CS PhD program at Georgia Tech in Fall 2024. His work spans adversarial machine learning, privacy, and agentic security, and he has published at top venues including CVPR, ICCV, and IEEE journals.
\vspace{5pt}

{\bf Fatih Ilhan} graduated from Bilkent University, Turkey, with BS and MSc in Computer Science (CS), and joined the CS PhD program in the Georgia Institute of Technology since 2021. Fatih's research interest lies in efficient AI and Machine Learning systems and algorithms, and published in IEEE and ACM journals, and top conferences, e.g., CVPR, ICDCS, NeurIPS, WWW.

\vspace{5pt}

{\bf Tiansheng Huang} graduated from Southern University, China, with BS and MS and started his CS PhD program in the Georgia Institute of Technology since 2022. He is working on safety alignment algorithms against harmful fine-tuning at user-level and published in IEEE and ACM journals, and top conferences, e.g., CVPR, ICML, ICLR, NeurIPS. 

\vspace{5pt}

{\bf Selim Tekin} graduated from Bilkent University, Turkey,  with BS and MSc in CS, and joined the CS PhD program in the Georgia Institute of Technology since 2022. Selim's PhD research interest lies in dynamic routing and ensemble learning for robust and high performance AI and ML systems, and has published in IEEE and ACM journals and top conferences, including ICML, EMNLP, CVPR, NeurIPS, WWW.

\vspace{5pt}

{\bf Sihao Hu} graduated with BS from Zhejiang University, China, MSc in Singapore National University, and joined the CS PhD program in the Georgia Institute of Technology since 2022. Sihao is working on Game AI agents and detection of  Fraudulent activities in Decentralized Financial and Crypto Systems, and has published in IEEE and ACM journals, and top conferences like WWW, CVPR, NeurIPS. 

\vspace{5pt}

{\bf Yichang Xu} graduated with BS from China Science and Technology University (Talented Class) in 2024 and joined the CS PhD program in the Georgia Institute of Technology since 2024. His research is centered on multi-modal agentic AI systems and algorithms. He has published in ACM journal and top conferences like CVPR and WWW.

\vspace{5pt}

{\bf Margaret L. Loper} is the Associate Director for Operations of the Information \& Communications Laboratory (ICL) at the Georgia Tech Research Institute.  Her research focus is modeling and simulation, specifically focused on parallel and distributed systems. She previously served as ICL’s Chief Scientist, where she led initiatives on Internet of Things, Information Security and Privacy and Smart Cities. Margaret holds a Ph.D. in CS from the Georgia Institute of Technology.

\vspace{5pt}

{\bf Ling Liu} is a Professor in the School of Computer Science at Georgia Institute of Technology. She directs the research programs in the Distributed Data Intensive Systems Lab (DiSL), examining various aspects of Internet-scale big data powered artificial intelligence (AI) systems, algorithms and analytics, including performance, reliability, privacy, security and trust. Ling’s current research is supported by National Science Foundation CISE programs, CISCO, and IBM.

\vfill

\end{document}